\documentclass{article}
\usepackage{spconf,amsmath,graphicx}
\usepackage{amsthm}
\usepackage[nolist,nohyperlinks]{acronym}
\usepackage{algorithm}
\usepackage{algpseudocode}
\usepackage{amssymb}
\usepackage{cite}
\newtheorem{lemma}{Lemma}
\usepackage{dsfont}
\newtheorem{theorem}{Theorem}
\renewenvironment{proof}[1][Proof]{\par\noindent\textbf{#1.} }{\hfill$\square$\par}
\usepackage{todonotes}  
\DeclareMathOperator{\Null}{Null}
\usepackage{bm} 
\newtheorem{assumption}{Assumption}
\usepackage{hyperref}
\usepackage{subcaption} 

\title{Multitask Learning with Learned Task Relationships}
\name{Zirui Wan and Stefan Vlaski\thanks{© 2025 IEEE. Personal use of this material is permitted. 
Permission from IEEE must be obtained for all other uses, 
in any current or future media, including reprinting/republishing 
this material for advertising or promotional purposes, creating new 
collective works, for resale or redistribution to servers or lists, 
or reuse of any copyrighted component of this work in other works.}}
  
\address{Department of Electrical and Electronic Engineering, Imperial College London, UK}

\begin{document}
\ninept

\maketitle

\begin{abstract}
Classical consensus-based strategies for federated and decentralized learning are statistically suboptimal in the presence of heterogeneous local data or task distributions. As a result, in recent years, there has been growing interest in multitask or personalized strategies, which allow individual agents to benefit from one another in pursuing locally optimal models without enforcing consensus. Existing strategies require either precise prior knowledge of the underlying task relationships or are fully non-parametric and instead rely on meta-learning or proximal constructions. In this work, we introduce an algorithmic framework that strikes a balance between these extremes. By modeling task relationships through a Gaussian Markov Random Field with an unknown precision matrix, we develop a strategy that jointly learns both the task relationships and the local models, allowing agents to self-organize in a way consistent with their individual data distributions. Our theoretical analysis quantifies the quality of the learned relationship, and our numerical experiments demonstrate its practical effectiveness.
\end{abstract}
\begin{keywords}
Decentralized multitask learning, graph signal processing, Gaussian Markov Random Field.
\end{keywords}
\section{Introduction}\label{sec:intro}
Classical federated learning and decentralized learning formulations often impose a consensus constraint—i.e., all agents must agree on a single global model or decision~\cite{Consensus1,Consensus2,Adaptation}. However, under heterogeneous local data or diverse task distributions, strict consensus can be statistically suboptimal because it enforces an overly restrictive compromise across tasks~\cite{Multitask1,Multitaks2}. This limitation has motivated \emph{multitask learning}, in which each agent maintains its own parameter vector while exploiting inter-task relationships through structural priors \cite{Multitask1,Multitaks2,stefan_reg1,stefan_reg2}. By explicitly modeling these relationships and incorporating them into the learning process, multitask learning can achieve superior estimation accuracy compared to consensus-based approaches.

In this work, we consider networked learning problems where each agent aims to estimate a parameter vector by minimizing its own individual cost:  
\vspace{-3mm}
\begin{equation}
\vspace{-1mm}
    w^o_k = \operatorname*{arg\,min}_{w_k}\;J_k(w_k).
\vspace{-1.5mm}
\end{equation}
\vspace{1pt}
When prior knowledge is available, it can be incorporated into multitask learning through a regularization term that augments the objective with a penalty enforcing desirable structure in the solution~\cite{regularization,stefan_reg1}. In particular, we consider: 
\vspace{-3mm}
\begin{equation}
\vspace{-0.8mm}
\operatorname*{arg\,min}_{{\scriptstyle \mathcal{W}}}
\;\mathcal{J}({\scriptstyle \mathcal{W}})
\;+\;
\frac{\eta}{2}\,{\scriptstyle \mathcal{W}}^\top \mathcal{R}\,{\scriptstyle \mathcal{W}}\label{eq:multitask_regularized},
\vspace{-2mm}
\end{equation}
where ${\scriptstyle \mathcal{W}}=\operatorname{col}\{w_{1},\ldots,w_{K}\}$  concatenates $w_k\in\mathbb{R}^M$ from $K$ agents into a column vector, 
$\mathcal{J}({\scriptstyle \mathcal{W}})\triangleq \sum_{k=1}^K J_k(w_k)$ is the aggregate cost, 
$\eta>0$ is a tuning parameter, and $\mathcal{R}$ is a positive semidefinite matrix.
In this work we adopt a \emph{graph smoothness} regularizer by setting 
$\mathcal{R}$ as a graph Laplacian matrix. This choice promotes similarity of the parameters across neighboring agents while still allowing heterogeneity.

Optimization with graph Laplacian regularization has been extensively studied. 
For example,~\cite{stefan_reg1,stefan_reg2,regularization,kernels} analyze iterative solutions and characterize how topology, stepsize, and regularization strength affect the steady-state performance. 
A key limitation of these works is their assumption of full knowledge of the Laplacian (both connectivity and edge weights), which is often unavailable in practice. 
Other efforts attempt to learn the Laplacian directly, typically from structured data over the graph~\cite{GMRF3,graphlearn}. In contrast, our work departs from these settings: the estimated Laplacian is not tied to data relationships but instead encodes latent \emph{task relationships} among agents, and must be inferred from noisy, non-cooperative parameter estimates, resulting in a coupled problem of inferring both task relationships and optimal parameters.

Another line of work adopts non-parametric strategies based on meta-learning~\cite{meta0,meta_learning1,meta_learning2} or proximal formulations~\cite{proximal1,proximal2}. Meta-learning methods adapt models across tasks by searching a common launch model. Proximal approaches, on the other hand, control personalization through penalties on the deviation from a reference model. While flexible and able to accommodate complex task structures, these methods typically require larger datasets and provide limited interpretability of the underlying task relationships.

Motivated by these challenges, we propose a strategy that jointly learns local models and their inter-task relationships. The dependencies among tasks are modeled through a Gaussian Markov Random Field (GMRF) whose \emph{unknown} precision matrix is constrained to the space of valid graph Laplacians and inferred from non-cooperative estimates of the local models. The estimated Laplacian is subsequently incorporated into the decentralized multitask learning procedure to promote structured cooperation among agents. We establish bounds on the Laplacian estimation error in the small-stepsize and high-dimensional regimes, showing an $O(\mu)$ dependence on the non-cooperative learning stepsize $\mu$. Finally, we evaluate the downstream learning performance when using the estimated Laplacian and compare it against several baseline methods. The proposed framework has potential applications in large-scale sensor networks, recommendation systems, and federated healthcare analytics~\cite{senser_networks,recommender_systems,health_care}, where agents must learn related but non-identical models while exploiting latent structural relationships.

Throughout the paper, all vectors are column vectors. Random quantities are in boldface; matrices are uppercase, and vectors/scalars are lowercase. $\otimes$ denotes the Kronecker product, $\operatorname{diag}(\cdot)$ constructs a block-diagonal matrix, and $\operatorname{vec}(\cdot)$ stacks the columns of a matrix into a vector. The notation $\|\cdot\|$ refers to the spectral norm for matrices and the $\ell_2$-norm for vectors.
\section{Gaussian Markov Random Field}
Assume that the true graph parameter ${\scriptstyle \mathcal{W}}^o=\operatorname{col}\{w^o_{1},\ldots,w^o_{K}\}$ is modeled by a Gaussian Markov Random Field (GMRF).  In this framework, each agent’s parameter $w_k$ is treated as a Gaussian random variable, with conditional dependence relations encoded by the edges of a connected, undirected graph. This construction captures the intuition that neighboring agents are more likely to have similar parameters, thereby promoting smoothness across the network. 

Each edge $(k,\ell)$ on the graph is assigned a nonnegative weight $a_{k\ell}$ reflecting the similarity between agents $k$ and $\ell$. Let $A$ denote the weighted adjacency matrix with entries $A_{k\ell}=a_{k\ell}$, and $D=\operatorname{diag}(d_1,\ldots,d_K)$ with $d_k=\sum_{\ell}a_{k\ell}$. The graph Laplacian is then defined as  
\begin{equation}
    L \triangleq D - A\label{eq:Laplacian}.
\end{equation}
The Laplacian is adopted as the precision matrix of the GMRF, directly tying the probabilistic model to the graph topology~\cite{GMRF1,GMRF2,GMRF3}. Accordingly, we state the following assumption:
\begin{assumption}[GMRF model]\label{assumption:GMRF}
The true parameter vector $\boldsymbol{{\scriptstyle \mathcal{W}^o}}$ is assumed to follow a multivariate Gaussian distribution:
\begin{equation}
     \boldsymbol{{\scriptstyle \mathcal{W}^o}} \sim \mathcal{N}\{0,\,\mathcal{L}^{\dagger}\}, \label{eq:GMRF}
\end{equation}
where $\mathcal{L} = L \otimes I_M$ and $(\cdot)^{\dagger}$ denotes the pseudo-inverse. 
Its probability density function is given by
\begin{equation}
f({\scriptstyle \mathcal{W}}^o) =
\frac{
\exp\!\left( -\tfrac{1}{2} (\boldsymbol{{\scriptstyle \mathcal{W}}}^o )^\top \mathcal{L}\, \boldsymbol{{\scriptstyle \mathcal{W}}}^o \right)
}{
\sqrt{\det\nolimits^*(2\pi \mathcal{L}^{\dagger})}
},
\end{equation}
where $\det^*(\cdot)$ denotes the pseudo-determinant and $\delta$ is the mean vector.
\end{assumption}
Since the Kronecker structure $\mathcal{L}=L \otimes I_M$ replicates the same graph-induced dependency across all feature dimensions, distribution~\eqref{eq:GMRF} also implies that every feature dimension provides an \emph{independent sample} of the dependency structure encoded by the graph Laplacian. As a result, the empirical covariance across features concentrates around $L^{\dagger}$, and the estimation error decays at the classical sub-Gaussian rate $O(K/M)$~\cite{High-dimensional_probability}.

\section{Multitask Learning Algorithm}
Under the GMRF prior \eqref{eq:GMRF}, the maximum a posteriori (MAP) estimate of 
${\scriptstyle \mathcal{W}^o}$ naturally takes the form of \eqref{eq:multitask_regularized}. 
The MAP estimator is given by \cite{smoothness,stefan_reg1}  
\begin{align}
        {\scriptstyle \mathcal{W}}_i^{*} 
        &= \operatorname*{arg\,min}_{{\scriptstyle \mathcal{W}}^o} 
        \Big\{-\log f_{\{\boldsymbol{x}_j\}_{j=1}^i|\boldsymbol{\mathcal{W}}^o}(\{x_j\}_{j=1}^i \mid {\scriptstyle \mathcal{W}^o}) 
        - \log f({\scriptstyle \mathcal{W}^o})\Big\}\label{eq:MAP}\\
        &= \operatorname*{arg\,min}_{{\scriptstyle \mathcal{W}}} \;
        \mathcal{Q}({\scriptstyle \mathcal{W}};\{x_j\}_{j=1}^i) 
        + \tfrac{1}{2} {\scriptstyle \mathcal{W}}^\top \mathcal{L}\,{\scriptstyle \mathcal{W}}\label{eq:map cost},
\end{align}
here, $\{x_j\}_{j=1}^i$ denotes the collection of data observed by all agents up to time $i$. 
By substituting the prior~\eqref{eq:GMRF} into~\eqref{eq:MAP} and defining the instantaneous loss as   
$\mathcal{Q}({\scriptstyle \mathcal{W}};\{x_j\}_{j=1}^i) \triangleq -\log f_{\{\boldsymbol{x}_j\}_{j=1}^i|\boldsymbol{\mathcal{W}}^o}(\{x_j\}_{j=1}^i \mid {\scriptstyle \mathcal{W}^o})$, we obtain the regularized multitask formulation \eqref{eq:map cost}.  

The cost function is then defined as the expected loss $\mathcal{J}({\scriptstyle \mathcal{W}}) \triangleq \mathbb{E}_{\boldsymbol{x}_j}\,
\mathcal{Q}({\scriptstyle \mathcal{W}};\boldsymbol{x}_j)$. This leads to the Laplacian-regularized optimization problem:
\begin{equation}
{\scriptstyle \mathcal{W}}^*
\;=\;
\operatorname*{arg\,min}_{{\scriptstyle \mathcal{W}}}
\;\mathcal{J}({\scriptstyle \mathcal{W}})
\;+\;
\frac{1}{2}\,{\scriptstyle \mathcal{W}}^\top \mathcal{L}\,{\scriptstyle \mathcal{W}}\label{eq:multitask cost},
\end{equation}
which matches the formulation in~\eqref{eq:multitask_regularized}.  

We are particularly interested in solving \eqref{eq:multitask cost} in the stochastic setting, where the data distribution—and hence the exact cost and gradient $\nabla \mathcal{J}({\scriptstyle \mathcal{W}})$—are unknown. In this case, agents implement a stochastic gradient descent recursion \cite{stefan_reg1,stefan_reg2}:  
\begin{equation}
    \boldsymbol{{\scriptstyle \mathcal{W}}}_i= (I_{MK}-\mu\mathcal{L})\boldsymbol{{\scriptstyle \mathcal{W}}}_{i-1}-\mu\widehat{\nabla \mathcal{J}}(\boldsymbol{{\scriptstyle \mathcal{W}}_{i-1}})\label{eq: DGD recursion},
\end{equation}
where $\mu>0$ is the stepsize and $\widehat{\nabla \mathcal{J}}({\scriptstyle \mathcal{W}})$ is a stochastic approximation of the gradient based on the available data. Due to the sparse structure of the graph Laplacian, the resulting algorithm is \emph{decentralized} by design. When $\mathcal{L}$ is known, this algorithm converges (for sufficiently small $\mu$) to the optimal MAP solution\cite{stefan_reg1}.

\section{Laplacian Estimation}
However, in practice, we cannot assume full knowledge of the Laplacian $L$. 
Instead, we rely on the structural prior in~\eqref{eq:GMRF} to estimate a suitable $\widehat{L}$. 
Since the true parameter vectors ${\scriptstyle \mathcal{W}^o}$ that encode inter-task relationships are not directly accessible, they must first be estimated locally in a non-cooperative manner, without knowledge of $L$. 
This is accomplished through a stochastic gradient descent recursion performed independently at each agent: 
\begin{equation}
    \boldsymbol{w}_{k,i} 
    = \boldsymbol{w}_{k,i-1}
    - \mu\,\widehat{\nabla J}_k(\boldsymbol{w}_{k,i-1})\label{eq:GD recursion},
\end{equation}
where each agent updates its own parameter estimate using only local data. The resulting estimates are then aggregated to approximate the covariance $\Sigma = L^\dagger$. 
Under the zero-mean model, we adopt the empirical covariance estimator:   
\begin{equation}
\widehat{\Sigma}
= \frac{1}{M}\,\operatorname{blktr}\!\big( (P\,{\scriptstyle \mathcal{W}}_i)(P\,{\scriptstyle \mathcal{W}}_i)^\top \big) \label{eq:covariance estimator},
\end{equation}
where $P \in \mathbb{R}^{KM\times MK}$ is the commutation matrix that reshapes ${\scriptstyle \mathcal{W}}_i$ into an element-wise stacking, 
and $\operatorname{blktr}(\cdot)$ denotes the block-trace operator that sums the $K\times K$ diagonal blocks. According to \eqref{eq:Laplacian}, $L$ is symmetric and positive semidefinite  with $\operatorname{rank}(L)=K-1$ and $\Null(L)=\Null(L^\dagger)=\mathrm{span}\{\mathds{1}\}$. 
In contrast, the empirical covariance estimate is full rank since additive noise fills the null space. 
To mitigate the noise amplification in the pseudo-inverse caused by rank mismatch \cite{perturbation_pseudo-inverses}, a subspace projection should be applied to $\widehat{\Sigma}$:
\begin{equation}
\widehat\Sigma^\perp \;\triangleq\; Q\,\widehat\Sigma\,Q, \; Q \;\triangleq\; I_K-\tfrac{1}{K}\mathds{1}\mathds{1}^\top\label{eq:subspace_projection}. 
\end{equation}

Finally, the Laplacian estimate is obtained as  
\begin{equation}
    \widehat{L} \;\triangleq\; \big(\widehat{\Sigma}^\perp\big)^\dagger,\; \widehat{\mathcal{L}}=\widehat{L} \otimes I_M.
    \label{eq:Lhat}
\end{equation}
The learned Laplacian is then incorporated into the decentralized multitask recursion:
\begin{equation}
    \widehat{\boldsymbol{{\scriptstyle \mathcal{W}}}}_i= (I_{MK}-\mu\widehat{\mathcal{L}})\widehat{\boldsymbol{{\scriptstyle \mathcal{W}}}}_{i-1}-\mu\widehat{\nabla \mathcal{J}}(\widehat{\boldsymbol{{\scriptstyle \mathcal{W}}}}_{i-1})\label{eq:DGD with learned L},
\end{equation}

\section{Estimation quality analysis}
Since \eqref{eq:GD recursion} produces noisy parameter estimates, 
we examine how this noise propagates into the covariance and Laplacian estimation 
by measuring the mean-squared-errors
$\mathbb{E}_{\boldsymbol{{\scriptstyle \mathcal{W}}}_i}\|\widehat{\Sigma}^\perp-L^{\dagger}\|^2$ 
and 
$\mathbb{E}_{\boldsymbol{{\scriptstyle \mathcal{W}}}_i}\|\widehat{L}-L\|$. These expectations characterize the effect of stochastic fluctuations in $\boldsymbol{{\scriptstyle \mathcal{W}}}_i$.
To establish this analysis, we introduce the following assumptions on the cost $J_k(\cdot)$ and the gradient noise process 
$\boldsymbol{{\scriptstyle \mathcal{S}}}_{i}(\cdot)$, defined as
\begin{equation}
    \boldsymbol{{\scriptstyle \mathcal{S}}}_{i}(\boldsymbol{{\scriptstyle \mathcal{W}}}_{i-1})
    = \widehat{\nabla \mathcal{J}}(\boldsymbol{{\scriptstyle \mathcal{W}}}_{i-1})
    - \nabla \mathcal{J}(\boldsymbol{{\scriptstyle \mathcal{W}}}_{i-1}).
\end{equation}
These assumptions are commonly satisfied by objective functions of interest in learning and adaptation, such as quadratic and logistic costs \cite{Adaptation}.

\begin{assumption}[Cost functions]\label{assumption:Risk function} Each individual cost $J_k(w_k)$ is assumed to be convex, twice differentiable, and with bounded Hessian satisfying:
\begin{equation}
     \nu I_M\leq\nabla^{2}J_k(w_k)\leq\delta I_M,
\end{equation}
where $0<\nu\leq\delta <\infty $.

2) The Hessian $\nabla^{2}J_k(\cdot)$ satisfies the Lipschitz condition for any $w_1$, $w_2\in\mathbb{R}^M$ and $k_H\geq0$:
\begin{equation}
    \|\nabla^{2}J_k(w_1)-\nabla^{2}J_k(w_2)\|\leq k_H\|w_1-w_2\|.
\end{equation}
\end{assumption} 

\begin{assumption}[Gradient noise]\label{assumption:Gradient noise} For any $\boldsymbol{{\scriptstyle \mathcal{W}}}_{i-1}$, gradient noise satisfies:
\begin{align}
    \mathbb{E}[\boldsymbol{{\scriptstyle \mathcal{S}}}_{i}(\boldsymbol{{\scriptstyle \mathcal{W}}}_{i-1})|\boldsymbol{{\scriptstyle \mathcal{W}}}_{i-1}] &= 0\\
     \mathbb{E}[\|\boldsymbol{{\scriptstyle \mathcal{S}}}_{i}(\boldsymbol{{\scriptstyle \mathcal{W}}}_{i-1})\|^4|\boldsymbol{{\scriptstyle \mathcal{W}}}_{i-1}] &\leq \beta^4\|{\scriptstyle \mathcal{W}}^o-\boldsymbol{{\scriptstyle \mathcal{W}}}_{i-1}\|+\sigma_s^4,
\end{align}
where $\beta,\, \sigma_s \geq0.$

2) The conditional covariance of $\boldsymbol{{\scriptstyle \mathcal{S}}}_{i}(\boldsymbol{{\scriptstyle \mathcal{W}}}_{i-1})$ defined as  
\[\mathcal{R}_{s,i}(\boldsymbol{{\scriptstyle \mathcal{W}}}_{i-1})\triangleq \mathbb{E}[\boldsymbol{{\scriptstyle \mathcal{S}}}_{i}(\boldsymbol{{\scriptstyle \mathcal{W}}}_{i-1})\boldsymbol{{\scriptstyle \mathcal{S}}}^\top_{i}(\boldsymbol{{\scriptstyle \mathcal{W}}}_{i-1})|\mathbb{F}_{i-1}]\] satisfies:
\begin{align}
        \|\mathcal{R}_{s,i}(\boldsymbol{{\scriptstyle \mathcal{W}}}_{i-1})-\mathcal{R}_{s,i}({\scriptstyle \mathcal{W}}^o)\| &\leq k_s\|\boldsymbol{{\scriptstyle \mathcal{W}}}_{i-1}-{\scriptstyle \mathcal{W}}^o\|^{\gamma_s}\\
         \mathcal{R}_s \triangleq \lim_{i \to \infty}\mathcal{R}_{s,i}({\scriptstyle \mathcal{W}}^o) &> 0,
\end{align}
where $k_s\geq 0$ and $0<\gamma_s\leq4$.
\end{assumption} 
Under the Assumption~\ref{assumption:Risk function} and~\ref{assumption:Gradient noise}, we can call on the following Theorem from ~\cite{probability1,probability2,Clustering}.
\begin{theorem}[Asymptotic Normality]\label{Theorem: Asymptotic normality}
For sufficiently small stepsize $\mu$ and as $i \to \infty$, 
the sequence generated by~\eqref{eq:GD recursion} converges in distribution to an approximately conditional Gaussian~\cite{probability1,probability2}:  
\begin{equation}
    \boldsymbol{{\scriptstyle \mathcal{W}}}_i \mid {\scriptstyle \mathcal{W}}^o 
    \;\sim\; \mathcal{N}({\scriptstyle \mathcal{W}}^o,\Pi),
\end{equation}
where $\Pi$ denotes the steady-state error covariance matrix, which depends on the realization of the true parameter ${\scriptstyle \mathcal{W}}^o$. 
In particular, $\Pi$ is the unique symmetric positive semidefinite solution to the discrete Lyapunov equation~\cite{Clustering}:  
\begin{align}
    U\Pi U-\Pi+\mu^2\mathcal{R}_s=&0\label{eq:discrete Lyapunov equation},\\
    U \triangleq I_{KM}-\mu\mathcal{H}, \; \mathcal{H} = \operatorname{diag}&(\nabla^{2}J_k(w^o_k)).
\end{align}
Moreover, $\Pi$ vanishes linearly with the stepsize~\cite{Clustering}:
\begin{equation}
\Pi = O(\mu) \label{eq:O(mu)}
\end{equation}
\end{theorem}
Theorem~\ref{Theorem: Asymptotic normality} shows that the parameter estimates become asymptotically Gaussian, regardless of the distribution of the underlying data. This Gaussian approximation enables a tractable analysis of the subsequent error propagation.


\begin{lemma}[Covariance estimation error] \label{lem:covariance error} Suppose Assumption~\ref{assumption:GMRF} through~\ref{assumption:Gradient noise} hold. 
For $M \gg K$, the projected sample covariance estimator in~\eqref{eq:covariance estimator} satisfies:
\begin{align}
    \mathbb{E}_{\boldsymbol{{\scriptstyle \mathcal{W}}}_i}\|\widehat{\Sigma}^\perp-L^{\dagger}\|^2&\leq \mathbb{E}_{\boldsymbol{{\scriptstyle \mathcal{W}^o}}}\bigl[
    \underbrace{c_1\bigr(\|\boldsymbol{\Phi}\|^2+\|L^\dagger\|^2\bigl)\bigl(\frac{K}{M}+\frac{K^2}{M^2}\bigr) }_{\text{covariance concentration}}\notag\\&\;\;+\, \underbrace{c_2\bigl(\operatorname{tr}(L^\dagger)\operatorname{tr}(\boldsymbol{\Phi})+\|\frac{1}{M}\operatorname{blktr}(\boldsymbol{\Phi})\|^2\bigr)}_{\text{bias}}\bigr]\label{covariance error bound},
\end{align}
where $c_1,c_2 \geq 0$ are nonnegative constants, and $\boldsymbol{\Phi}\triangleq P\boldsymbol{\Pi}P^\top$, with $P$ denoting the commutation matrix. 
\end{lemma}
\begin{proof}
    Omitted due to space limitations.
\end{proof}
Since $\boldsymbol{\Phi}$ is a permutation of the steady-state error covariance $\boldsymbol{\Pi}$, it also depends on the true parameter realization ${\scriptstyle \mathcal{W}^o}$. The additional expectation $\mathbb{E}_{\boldsymbol{{\scriptstyle \mathcal{W}^o}}}[\cdot]$ on the right-hand side of \eqref{covariance error bound} accounts for this dependency, ensuring that the bound in Lemma~\ref{lem:covariance error} holds uniformly over both the non-cooperative estimates ${\boldsymbol{{\scriptstyle \mathcal{W}}}_i}$ and ture parameters $\boldsymbol{{\scriptstyle \mathcal{W}}^o}$.

The error bound in Lemma~\ref{lem:covariance error} offers useful insights into the quality of covariance estimation. 
In particular, the estimator $\widehat{\Sigma}^\perp$ is not consistent for finite stepsizes $\mu$: even as the number of samples $M \to \infty$, certain terms on the right-hand side remain non-vanishing, leaving a nonzero \emph{bias} in the limit.  
More precisely, the first term in~\eqref{covariance error bound} decays at rate $O\!\left(\tfrac{K}{M}\right)$ due to covariance concentration~\cite{High-dimensional_probability}, while the last term persists and contributes to the asymptotic bias:
\begin{equation}
     \lim_{M \to \infty} \mathbb{E}_{\boldsymbol{{\scriptstyle \mathcal{W}}}_i}\|\widehat{\Sigma}^\perp-L^{\dagger}\|^2  = c_2\bigl(\operatorname{tr}(L^\dagger)\mathbb{E}_{\boldsymbol{{\scriptstyle \mathcal{W}^o}}}[\operatorname{tr}(\boldsymbol{\Phi})]+\mathbb{E}_{\boldsymbol{{\scriptstyle \mathcal{W}^o}}}\|\frac{1}{M}\operatorname{blktr}(\boldsymbol{\Phi})\|^2\bigr)
    \label{bias}.
\end{equation}
Since $\boldsymbol{\Pi}$ vanishes with the stepsize as in~\eqref{eq:O(mu)}, it follows that $\boldsymbol{\Phi}=O(\mu)$. 
Consequently, the bias term decreases with the stepsize at rate $O(\mu)$, and thus:
\begin{equation}
    \lim_{\mu \to 0}\;\lim_{M \to \infty}\, \mathbb{E}_{\boldsymbol{{\scriptstyle \mathcal{W}}}_i}\|\widehat{\Sigma}^\perp-L^{\dagger}\|^2 = 0 \label{eq:zero bias}.
\end{equation}
\begin{theorem}[Laplacian estimation error]\label{Theorem: Laplacian estimation error} Suppose the conditions of Lemma~\ref{lem:covariance error} hold. By combining the covariance error bound with the pseudo-inverse perturbation results in Theorem~4.1 of~\cite{perturbation_pseudo-inverses}, 
we obtain that, for sufficiently small stepsize $\mu$ and sufficiently large sample size $M$:
    \begin{align}
        \mathbb{E}_{\boldsymbol{{\scriptstyle \mathcal{W}}}_i}\|\widehat L-L\|^2 &\overset{\text{\cite{perturbation_pseudo-inverses}}}{\leq}\,c_3\|L\|^2\,\mathbb{E}_{\boldsymbol{{\scriptstyle \mathcal{W}}}_i}\|(\widehat{\Sigma}^{\perp})^{\dagger}\|^2\,\mathbb{E}_{\boldsymbol{{\scriptstyle \mathcal{W}}}_i}\|\widehat{\Sigma}^{\perp}-L^{\dagger}\|^2\\
        &= O(\mu)\label{L error bound}.
    \end{align}
In particular, analogous to~\eqref{eq:zero bias}, the asymptotic bias vanishes in the joint limit:  
    \begin{equation}
    \lim_{\mu \to 0}\;\lim_{M \to \infty}\, \mathbb{E}_{\boldsymbol{{\scriptstyle \mathcal{W}}}_i}\|\widehat L-L\|^2 = 0 \label{eq:zero L bias}.
\end{equation}
\end{theorem}
\begin{proof}
    Omitted due to space limitations.
\end{proof}
From the bounds in~\eqref{covariance error bound}--\eqref{eq:zero L bias}, 
we conclude that with sufficiently large $M$ and sufficiently small stepsize $\mu$, 
the Laplacian estimation error decays at order $O(\mu)$ and vanishes in the joint limit $\mu \to 0$, $M \to \infty$. At the same time, the dependence on $\operatorname{tr}(L^{\dagger})$ and $\|L\|^2$ highlights the role of graph structure.
$\operatorname{tr}(L^{\dagger})$ is the sum of the inverses of the nonzero Laplacian eigenvalues. 
It becomes large when the graph is weakly connected, since small nonzero eigenvalues inflate the sum. 
In contrast, $\|L\|^2$ is controlled by the largest node degree and edge weights, and therefore increases in graphs with high-degree hubs or heavily weighted edges.
Thus, graphs with poor connectivity or highly unbalanced structure amplify estimation errors, 
making them intrinsically more difficult to recover accurately.

\section{Simulation Results}
\begin{figure}[t]
    \centering
    \includegraphics[width=0.85\linewidth]{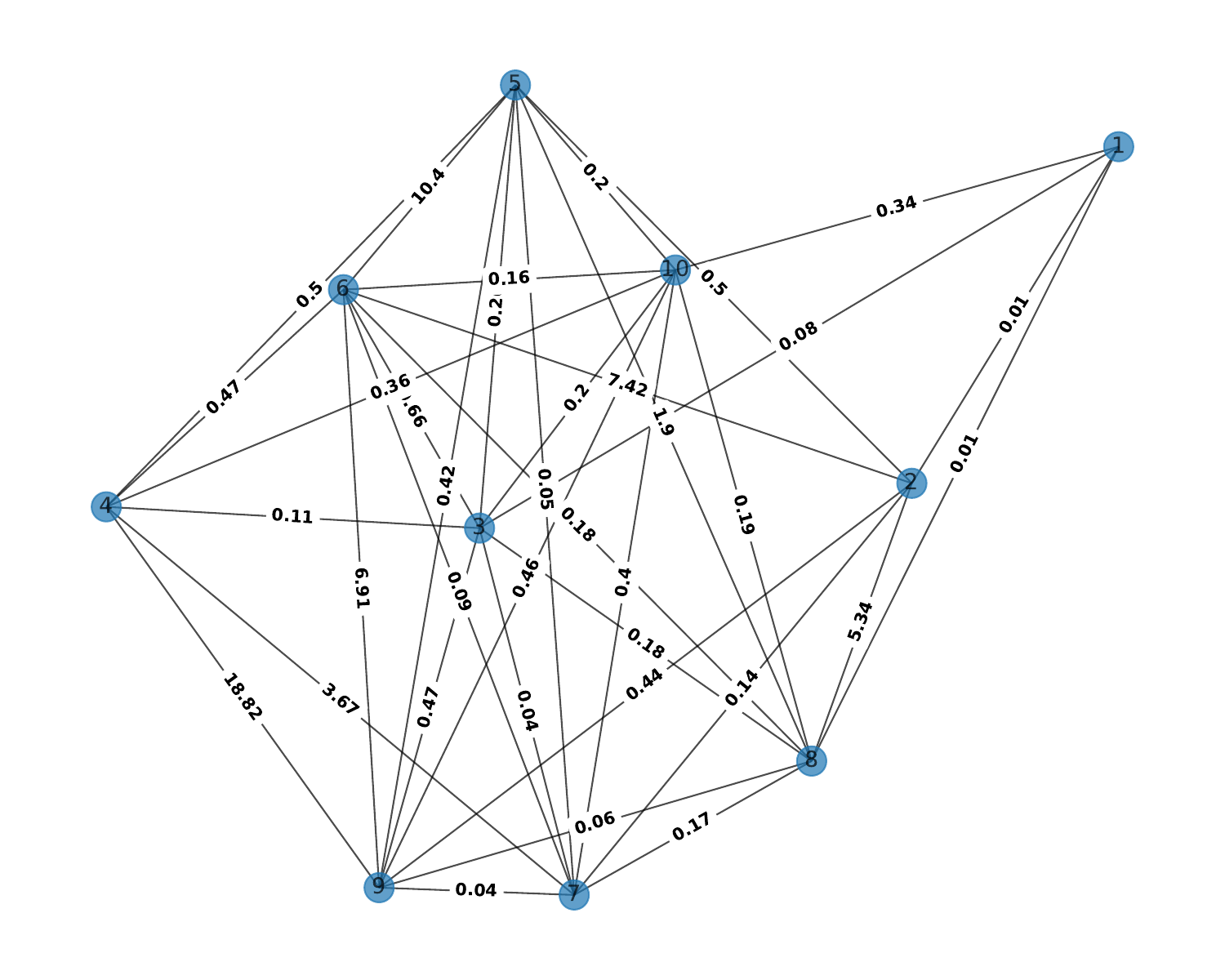}
    \vspace{-2mm}
    \caption{Graph topology with assigned edge weights.}
    \label{fig:graph}
\end{figure}

\begin{figure}[t]
    \centering
    \includegraphics[width=0.8\linewidth]{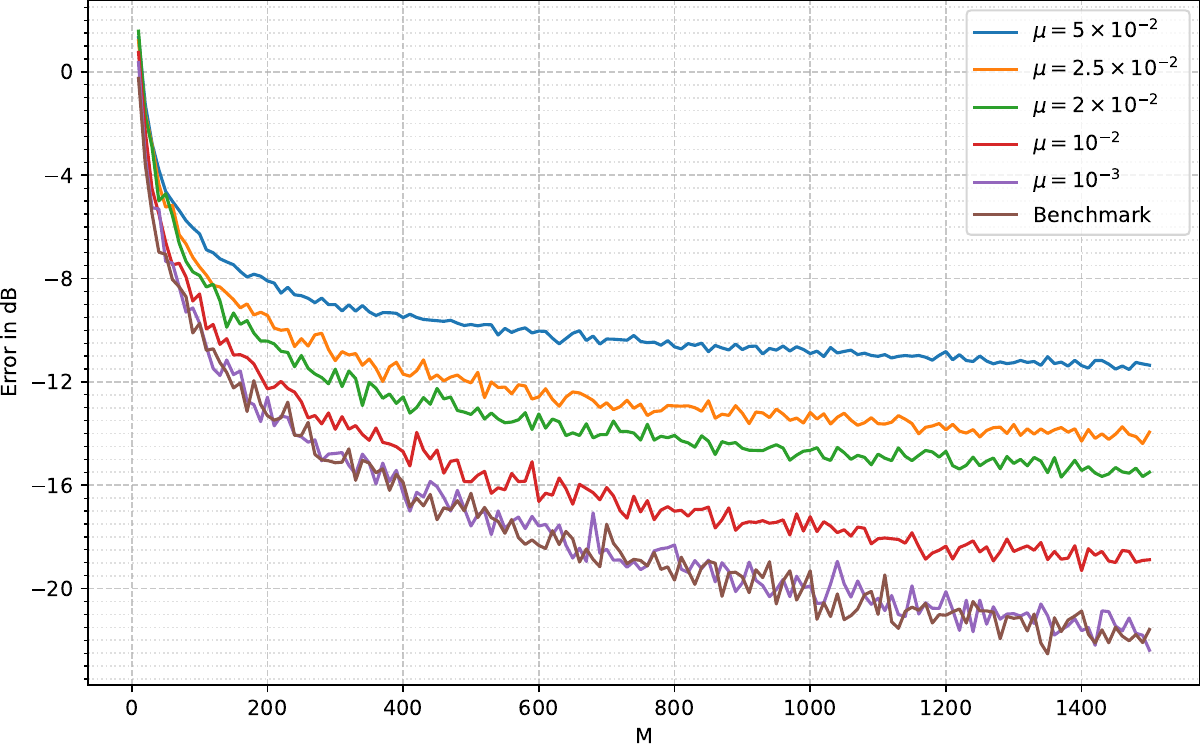}
    \vspace{-4mm}
    \caption{Covariance estimation error $\|\widehat{\Sigma}^{\perp}-L^{\dagger}\|^2$ versus $M$.}
    \label{fig:fig1}
    \vspace{-4mm}
\end{figure}

\begin{figure}[t]
    \centering
    \includegraphics[width=0.8\linewidth]{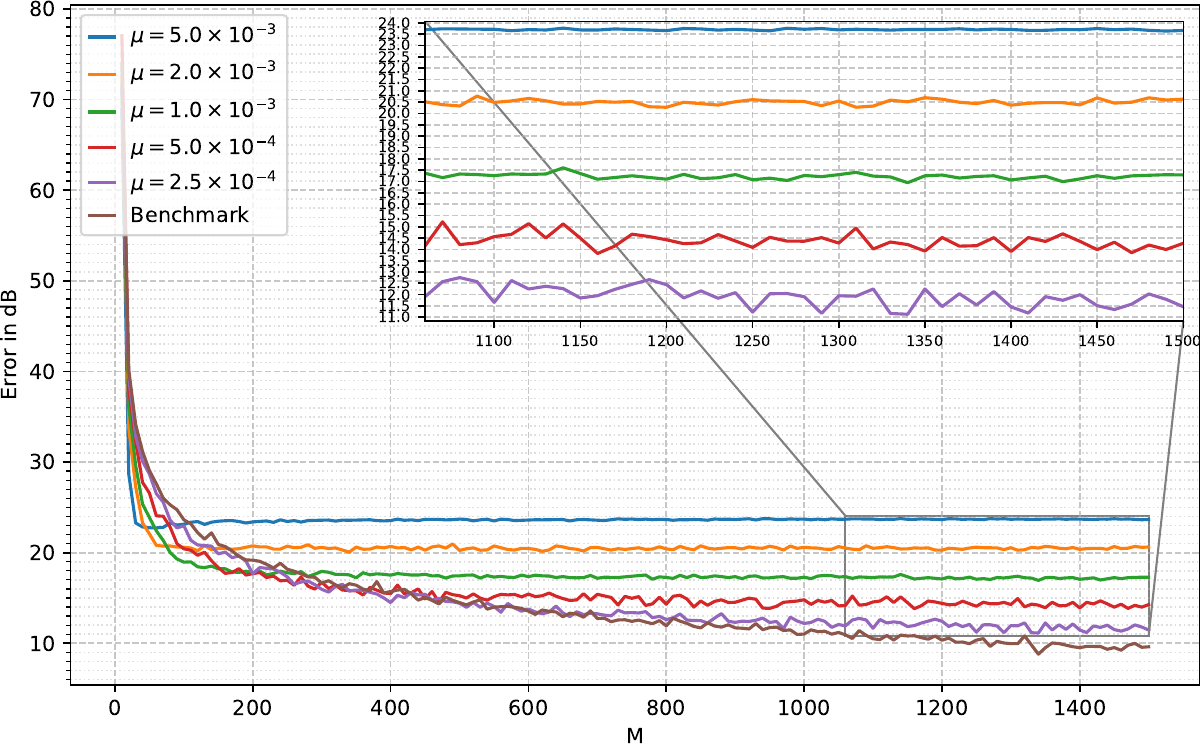}
    \vspace{-2mm}
    \caption{Laplacian estimation error $\|\widehat L-L\|^2$ versus $M$.}
    \label{fig:fig2}
\end{figure}

We construct a connected, undirected network with $K=10$ agents and maximum node degree $8$ shown in Figure~\ref{fig:graph}. To create an unbalanced weighted topology that better reflects heterogeneous task relationships and provides a more challenging test for the multitask learning algorithm, each edge weight is drawn from a mixture of two uniform distributions: with probability $0.3$ from $\operatorname{unif}(1,20)$ (large weights), and otherwise from $\operatorname{unif}(0,0.5)$ (small weights). 

Each agent is subjected to streaming data $\{\boldsymbol{d}_k(i), \boldsymbol{u}_{k,i}\}$, satisfying a linear regression model:
\begin{equation}
    \boldsymbol{d}_k(i) = \boldsymbol{u}^\top_{k,i}w^o_k + \boldsymbol{v}_k(i),\; k=1,...,K.
\end{equation}
The processes $\{\boldsymbol{u}_{k,i}, \boldsymbol{v}_k(i)\}$ are zero-mean jointly wide-sense stationary with: i) $\mathbb{E}\boldsymbol{u}_{k,i}\boldsymbol{u}^\top_{\ell,i}=\sigma^2_{u,k}I_M$ if $k=\ell$ and zero otherwise; ii) $\mathbb{E}\boldsymbol{v}_{k}(i)\boldsymbol{v}_{\ell}(i)=\sigma^2_{v,k}$ if $k=\ell$ and zero otherwise; iii) $\boldsymbol{u}_{k,i}$ and $\boldsymbol{v}_k(i)$ are independent. According to \eqref{eq:MAP}, the cost functions take the mean-square-error form:

\begin{equation}
    J_k(w_k) = \frac{1}{2}\mathbb{E}|{d}_k(i)-\boldsymbol{u}^\top_{k,i}w_k|^2.
\end{equation}
We run algorithm~\eqref{eq:GD recursion} with different stepsizes 
$\mu \in \{5\times 10^{-2},\,2.5\times 10^{-2},\,2\times 10^{-2},\,10^{-2},\,10^{-3}\}$ until convergence, 
and the resulting estimates $\scriptstyle \mathcal{W}_i$ are used in \eqref{eq:covariance estimator} and \eqref{eq:subspace_projection} to approximate 
the true covariance $\Sigma$. These estimates are compared against a benchmark constructed from the true 
${\scriptstyle \mathcal{W}}^o$.
Figure~\ref{fig:fig1} provides a numerical verification of the bound in 
\eqref{covariance error bound}. The $y$-axis reports the covariance estimation error 
$\|\widehat{\Sigma}^{\perp}-L^{\dagger}\|^2$, averaged over $100$ Monte-Carlo trials of 
${{\scriptstyle \mathcal{W}}}_i$ and $100$ realizations of ${{\scriptstyle \mathcal{W}}}^o$ drawn from the GMRF model~\eqref{eq:GMRF}
to approximate the expectation.  
We observe that when $M$ is sufficiently large, the estimation error converges 
to the bias terms characterized in \eqref{bias}, scaling consistently with the 
theoretical trend $O(\mu)$. This trend is evident in Figure~\ref{fig:fig1}; for instance, at $M=1500$, 
halving the stepsize $\mu$ reduces the error magnitude by a factor of $1/2$, 
which corresponds to a decrease of approximately $-3$ dB in the error curve.
Conversely, when $\mu$ is very small and $M$ is not sufficiently large, the error 
is dominated by the covariance concentration terms, which scale as 
$O\!\left(\tfrac{K}{M}\right)$. 
This explains why the curve corresponding to $\mu=10^{-3}$ nearly overlaps with the benchmark.

Figure~\ref{fig:fig2} presents a numerical verification of the bound in~\eqref{L error bound}. 
Laplacian estimation results with $\mu \in \{5\times 10^{-3},\,2\times 10^{-3},\,10^{-3},\,5\times 10^{-4},\,2.5\times 10^{-4}\}$ 
are compared against the benchmark. 
Similar to Figure~\ref{fig:fig1}, the Laplacian estimation errors 
$\|\widehat{L}-L\|^2$ are averaged over ${{\scriptstyle \mathcal{W}}}_i$ and ${{\scriptstyle \mathcal{W}}}^o$. 
We observe that when the stepsize $\mu$ is sufficiently small, the steady-state error 
also follows the theoretical trend $O(\mu)$.

\begin{figure}[t]
    \centering
    \includegraphics[width=0.8\linewidth]{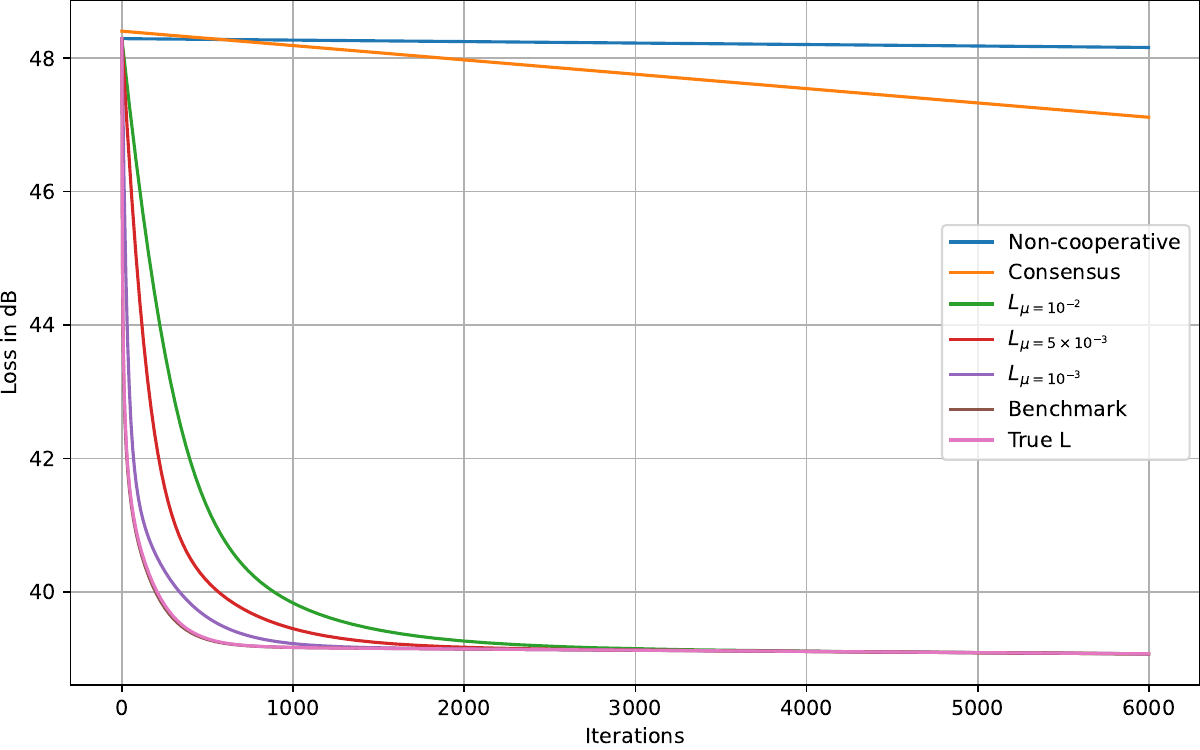}
    \vspace{-2mm}
    \caption{Learning performance of different algorithms.}
    \label{fig:fig3}
\end{figure}
Finally, Figure~\ref{fig:fig3} shows the transient learning performance, measured by the mean-squared deviation $\frac{1}{K}\mathbb{E}\|{\scriptstyle \mathcal{W}^o}-\boldsymbol{{\scriptstyle \mathcal{W}}}_i\|^2$, for different algorithms under the same adaptation stepsize $2\times 10^{-2}$ and $M = 1500$. The comparison includes the non-cooperative recursion~\eqref{eq:GD recursion}, the consensus strategy~\cite{Adaptation}, and the multitask recursion~\eqref{eq: DGD recursion} with Laplacians of varying estimation accuracy. The results show that the proposed multitask strategy generally learns faster by leveraging the estimated Laplacian to coordinate updates among statistically related agents, and that higher-quality Laplacian estimates yield correspondingly better performance.

\section{Conclusion}

In this work, we proposed a distributed multitask learning framework that jointly estimates local models and the underlying task relationships. By modeling inter-task dependencies through a GMRF model, we derived a practical strategy that learns the graph Laplacian from non-cooperative estimates. Our theoretical analysis established bounds on the Laplacian estimation error, highlighting their dependence on the stepsize and feature dimensions. Simulation results validated these bounds and further demonstrated that the learned Laplacian enables faster and more effective adaptation compared to non-cooperative and consensus strategies.




\newpage


\bibliographystyle{IEEEbib}
\bibliography{main}

\end{document}